\pdfoutput=1

\documentclass[11pt]{article}

\usepackage[numbers]{natbib}

\usepackage{booktabs}       
\usepackage{amsfonts}       
\usepackage{amsmath}
\usepackage{graphicx}
\usepackage{amsthm}
\usepackage{amssymb}
\usepackage{multirow}
\usepackage{makecell}
\usepackage[vlined,linesnumbered,ruled,resetcount]{algorithm2e}
\usepackage[colorlinks,linkcolor=magenta,filecolor=blue,citecolor=blue,urlcolor=blue]{hyperref}%
\usepackage[top=1in, left=1in, right=1in, bottom=1in]{geometry}
\usepackage{subfigure}
\usepackage{amssymb}
\usepackage{pifont}
\usepackage{mathtools}
\usepackage{stmaryrd}
\usepackage[T1]{fontenc}

\usepackage{authblk}
\usepackage{enumitem}
\usepackage{algorithmic}

\usepackage{graphicx}
\usepackage{booktabs}
\usepackage{colortbl}
\usepackage{url}
\usepackage{amsmath}
\usepackage{caption}
\usepackage{subcaption}
\usepackage{xcolor}

\theoremstyle{definition}

\newcommand{\wh}{\widehat}
\newcommand{\wt}{\widetilde}

\renewcommand{\epsilon}{\varepsilon}
\renewcommand{\phi}{\varphi}

\renewcommand{\tilde}{\wt}
\renewcommand{\hat}{\wh}

\newcommand{\sentref}{x}
\newcommand{\senthyp}{\hat{x}}

\newcommand{\vref}{\mathbf{x}}
\newcommand{\vhyp}{\mathbf{\hat{x}}}

\makeatletter
\newcommand*{\RN}[1]{\expandafter\@slowromancap\romannumeral #1@}
\makeatother

\makeatletter
\newcommand{\printfnsymbol}[1]{%
  \textsuperscript{\@fnsymbol{#1}}%
}
\makeatother

\title{TensorOpera Router: A Multi-Model Router \\ for Efficient LLM Inference}

\author[1]{Dimitris Stripelis}
\author[1]{Zijian Hu}
\author[1]{Jipeng Zhang}
\author[1]{Zhaozhuo Xu}
\author[1]{Alay Dilipbhai Shah}
\author[1]{Han Jin}
\author[1]{Yuhang Yao}
\author[1]{Salman Avestimehr}
\author[1]{Chaoyang He}
\affil[1]{TensorOpera, Inc., Palo Alto, CA, USA}
\affil[ ]{{\texttt{contact@tensoropera.com}}}
\begin{document}

\maketitle
\begin{abstract}
    With the rapid growth of Large Language Models (LLMs) across various domains, numerous new LLMs have emerged, each possessing domain-specific expertise. This proliferation has highlighted the need for quick, high-quality, and cost-effective LLM query response methods. Yet, no single LLM exists to efficiently balance this trilemma. Some models are powerful but extremely costly, while others are fast and inexpensive but qualitatively inferior. To address this challenge, we present \textit{TO-Router}, a non-monolithic LLM querying system that seamlessly integrates various LLM experts into a single query interface and dynamically routes incoming queries to the most high-performant expert based on query's requirements. Through extensive experiments, we demonstrate that when compared to standalone expert models, \textit{TO-Router} improves query efficiency by up to 40\%, and leads to significant cost reductions of up to 30\%, while maintaining or enhancing model performance by up to 10\%.

\end{abstract}

\section{Introduction}
Large Language Models (LLMs) have demonstrated remarkable performance across a diverse set of challenging domain-specific tasks~\cite{beeching2023open}. However, no single LLM can outperform all others across every task and use case~\cite{shnitzer2023large}. Recent works~\cite{hu2024routerbench,ong2024routellm,ding2024hybrid} highlight the urgent need for efficient tools that can unify the expertise of multiple LLMs, combining them into a single cohesive unit. Given the increased costs and latency associated with querying models hosted at different providers~\cite{chen2023frugalgpt}, it is critical for multi-LLM querying systems to efficiently route queries to the most appropriate LLM expert. This must be done while balancing query execution throughput, monetary costs, and model performance—a challenge we term the multi-LLM routing trilemma.

Our aim is to provide an empirical solution to this trilemma by showcasing the potential of a multi-LLM routing system that improves this balance. We propose an LLM routing system, called \textit{TensorOpera-Router} (hereinafter referred to as \textit{TO-Router}), to explore the feasibility of building a multi-LLM routing model that leverages the collective power of multiple LLM experts. \textit{TO-Router} aims to efficiently, inexpensively, and accurately answer query prompts by selecting the most cost-effective and suitable LLM from a diverse set of expert models. Our contributions are as follows:
\begin{itemize}
\setlength{\itemsep}{1pt}
\setlength{\parskip}{0pt}
\setlength{\parsep}{0pt}
\item We empirically demonstrate the promise of different routing methods developed through the \textit{TO-Router} system in balancing query execution time, query cost, and model performance, leading to significant gains.
\item We show that, on average, our routing system outperforms standalone model experts.
\item We demonstrate that routing methods trained to learn the embedding query space outperform naive routing methods.
\item We introduce a soft label based approach based on BERT similarity score. to train the proposed model routing methods.
\item We present a routing method based on a pre-trained BERT model that exhibits the best performance.
\end{itemize}

\begin{figure*}
    \centering
    \includegraphics[width=\linewidth]{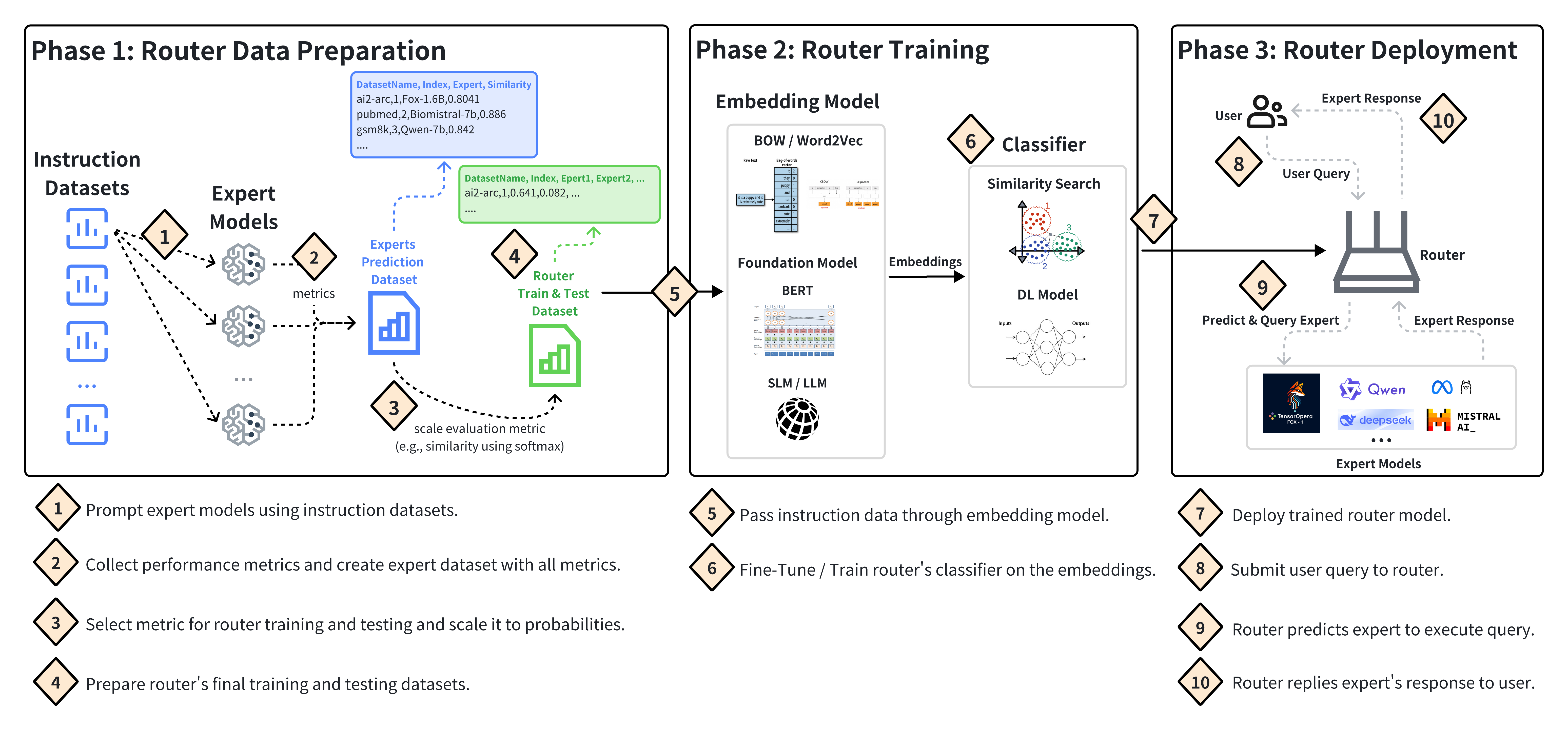}
    \caption{TO-Router system's overview of router data preparation, router model training and deployment pipelines.}
    \label{fig:router_pipeline}
\end{figure*}

\section{Background \& Related Work}

\textbf{LLM Routing.} Depending on the mechanism used by routing methods to decide the most suitable LLM(s) to answer a given prompt, two distinct routing categories have been recently introduced: \textit{predictive/classification routers}, which do not generate LLM outputs in advance, but instead, they predict the best LLM to handle a given prompt based on specific performance metrics~\cite{hu2024routerbench,ong2024routellm,srivatsa2024harnessing} and \textit{cascading routers}, which refer to routing methods that process a query request by executing it over a series or combinations of LLMs~\cite{chen2023frugalgpt} until specific quality criteria are met. To train the predictive routers, different training methods have been recently introduced that leverage data augmentation techniques and human preference data~\cite{ong2024routellm} or existing benchmark datasets~\cite{shnitzer2023large} to improve routing predictions. In this work, we too develop and evaluate predictive routing methods trained on standardized benchmark datasets to efficiently classify and direct query prompts to the best LLM expert.

\textbf{Mixture-of-Experts.} A typical MoE architecture~\cite{jordan1994hierarchical} consists of a set of expert models trained to specialize in different data regions and a gating network model that determines the contribution of each expert to the final prediction. Recently, MoEs have witnessed wide adoption in the LLM domain as well, where multiple MLP experts are integrated into encoder and decoder blocks, to boost the training of extremely large networks~\cite{shazeer2017outrageously,jiang2024mixtral,fedus2022switch}. Similar to these MoE approaches, the LLM routing methods can be seen a special case of an MoE architecture, where the predictive routing model is the gating mechanism and the pool of LLMs the set of available experts.

\textbf{Ensemble Learning.} Model routing also bears similarities with ensemble machine learning~\cite{zhou2012ensemble} techniques that seek to provide better predictive outcomes by combining the predictions of multiple models. A key distinction between routing and ensemble techniques, like bagging~\cite{breiman1996bagging}, and boosting~\cite{freund1997decision}, is that models participating in an ensemble are typically trained on the (whole or subsets of) same dataset and therefore assumed to have a similar expertise. However, the router predicts and retrieves the predictions out of a varying set of LLMs experts that have been trained on highly diverse sets of data distributions.

\section{TensorOpera Router System Overview}
To effectively learn and deploy a multi-LLM routing model, a sequence of different critical development phases need to be executed, from data preparation to router model training and evaluation and model deployment/serving. The proposed TO-Router system's end-to-end pipeline shown in Figure~\ref{fig:router_pipeline} facilitates the development of these phases and in practice has drastically helped to swiftly develop, prototype and deploy different model routing methods into real-world settings.~\footnote{Source code to be released.}

\textbf{Phase 1: Router Data Preparation.} The generation of the training and testing dataset for the routing model is a multi-step process. First, we need to find the appropriate domain specific (e.g., bio, coding, physical sciences) instruction datasets and model experts to which we want the routing model to learn propagating relevant query prompts. Thereafter, we perform a forward pass over each expert model (step 1) to collect the associated metrics required to train and test the performance of the routing model and create the experts prediction dataset (step 2). In this work, we collect the following metrics per instruction prompt: \{negative log likelihood, BERT similarity score (BERTSim), inference time in seconds, total input tokens, total output tokens\}; for more details on these metrics, please see section~\ref{subsec:evaluation_criteria}. Once the expert prediction dataset is created, we select one of the collected metrics to generate soft labels (step 3) and prepare the final training and testing dataset for the routing model (step 4). In the current work, we use the BERTSim scores to create soft labels and train the routing expert model classifier. We use soft labels, since we want the routing model to learn the ranking of the experts in terms of their prediction performance. To generate the soft labels of each expert model and for each instruction record, we pass the selected metric (e.g., similarity score, log loss), through a softmax function with temperature. For instance, for the $r$-instruction record, the expert (class) softmax probability $\phi_r$ is given by: $\phi_r(\mathbf{x}; T) = \tfrac{\exp\left(\tfrac{x_i}{T}\right)}{\sum_{j=1}^E \exp\left(\tfrac{x_j}{T}\right)}$, where $E$ is the total number of experts, $T$ is the temperature value, and $\mathbf{x} = (x_1, x_2, \ldots, x_E)$ is the vector of metric scores. In our evaluation, we generate expert's soft labels based on the BERT similarity scores and with a temperature value of $T = 10$.

\textbf{Phase 2: Router Training.} Once the router's training and testing dataset is created, we pass the instruction records through the router's embedding model, e.g., Bag-of-Words, TF-IDF, BERT or other small or large language models, to create their vectorized representation (step 5). Then, we use the generated embeddings to train the prompt-to-expert classifier (step 6), using non-parametric, supervised learning approaches (e.g., kNN), classical deep learning models (e.g., MLP) or more advanced language sequencing pre-trained models (e.g., BERT). We provide more information on these routing models in section~\ref{subsec:routing_methods}.

\textbf{Phase 3: Router Deployment.} When the final routing model is trained, the model is deployed as a standalone endpoint on the platform (step 7), ready to receive user queries (either through CLI or web interface). Whenever a new user query is submitted, the router first tokenizes and encodes the text of the incoming query prompt using the tuned embedding model from Phase 2 (step 8). Subsequently, the router performs a forward pass over the trained/fine-tuned classification model (e.g., MLP, BERT) and predicts the most relevant expert model (step 9). Depending on which expert model the classification model predicts, the router selects the respective expert-prompt adaptor to submit and execute the query. Once query execution completes, the router receives the reply from the expert model and forwards it back to the end user (step 10). Throughout the router's deployment time, the platform provides the necessary monitoring capabilities to troubleshoot and tune the routing model, such as number of requests, queries' semantic context, expert models hitting frequency, and total costs.

\section{Experiments}\label{sec:experiments}
In this section we discuss the metrics, expert models, benchmark datasets and routing methods we considered for evaluating the TO-Router system.  

\subsection{Evaluation Criteria}\label{subsec:evaluation_criteria}
All expert models and routing methods are evaluated on four dimensions: (1) total inference cost, (2) throughput, (3) BERT similarity score, and (4) negative log loss (NLL).

\textbf{Total Inference Cost.} For any expert model the total cost to execute a given test query is measured based on the input and output token costs. For a model $m$ that was prompted with a sequence of test queries that were used a total number of $T_i$ input tokens, and the model generated a total number of $T_o$ output tokens, with a $c_i$ and $c_o$ cost per 1 million input and output tokens, respectively, the total cost for the entire test query sequence is measured by: $
C_m = \tfrac{T_{i}}{1e6} * c_{i} + \tfrac{T_{o}}{1e6} * c_{o}
$.
In the case of the routing methods that did not use one single model to answer the sequence of testing queries but routed different testing queries to different expert models $M$, the total cost is measured as: $C_r = \sum_{m \in M} C_m$. To measure the querying of standalone deployed expert models, we handpicked the price per million input and output tokens from different model providers. We provide the input and output token costs per model architecture in Table~\ref{tbl:model_pricing} in the Appendix section.

\textbf{Throughput.} To measure the querying execution performance of a expert model and of different routing methods for the entire test query set, we compute the throughput for each query as the fraction of total output tokens $T^o_m$, generated by each model $m$, over the inference time in seconds, i.e., time from query submission to query completion, $t^s_m$. Specifically, the throughout for a single test query $i$ is measured as $\tau_i = \tfrac{T^o_m}{t^s_m}$. For the entire set of test queries $N$, the mean throughput $\tilde{\tau}$ is computed as: $\tilde{\tau} = \frac{1}{N} \sum_{i}^N \tau_i$.

\textbf{BERTSim.} Given that each expert model uses its own vocabulary and tokenizer and to ensure that there is an equitable comparison between the responses generated by each expert, we evaluate the vectorized text similarity between the ground truth and the predicted answer of an expert through the cosine distance on the BERT embeddings. Such a vector representation allows for a soft measure of similarity~\cite{zhang2019bertscore}. We refer to this similarity score as BERTSim~\cite{zhang2019bertscore}. The cosine similarity of a reference (ground truth) vector $\sentref_i$ and a candidate (predicted) vector $\senthyp_j$ is computed as: $\frac{\vref_i^\top \vhyp_j}{\|\vref_i\| \|\vhyp_j\|}$. For every expert model and routing method we measure the BERTSim score across all test queries and we compute the final BERTSim score as the mean of all scores.

\textbf{Negative Log-Likelihood.} We use the Negative Log-Likelihood (NLL) to measure the quality of the probabilistic predictions made by each expert model. Lower NLL values are indication that the model is assigning higher probabilities to the true classes and therefore reflecting better performance. In principle, a single sequence's NLL is defined as:
\[
\mathcal{L}_{\text{NLL}} = - \sum_{t=1}^{T} \log P(y_t \mid X, y_{1:t-1})
\]
where \( P(y_t \mid X, y_{1:t-1}) \) is the predicted probability of the \( t \)-th token in the sequence given the input sequence \( X \) and the previous tokens \( y_{1:t-1} \). In our evaluation, we measure the mean NLL over the generated sequence of every expert model and routing method across all test queries.

\subsection{Expert Models}
We choose several representative models across different domains as the expert models to verify the effectiveness of our routing method in the TO-Router system. For the Biomedical domain, we selected two variants from Llama-3-8B (\textbf{BioLlama-7B})~\citep{shao2024deepseekmath} and Mistral-7B (\textbf{BioMistral-7B})~\citep{labrak2024biomistral} models~\footnote{In our evaluation, we refer to each model using its name in bold fonts.}. Both models achieve excellent performance across many biomedical evaluation benchmarks. In the code domain, we select Meta's officially released Llama2-7B (\textbf{CodeLlama-7B})~\citep{roziere2023codellama} variant trained on code datasets. In the general instruction-following domain, we incorporate three instruction-tuned versions of LLMs across different sizes, i.e., \textbf{Fox-1.6B} a recently introduced powerful small language model, Mistral-7B-Instruct (\textbf{MistralAI-7B})~\citep{jiang2023mistral}, and Qwen-7B-Instruct (\textbf{Qwen-7B})~\citep{yang2024qwen2technicalreport}. Finally, for the math domain, we choose a strong reasoning model trained on large amounts of math documents, MathDeepSeek-7B-Instruct (\textbf{MathDeepSeek-7B})~\citep{guo2024deepseekcoder}. For more details regarding models architecture and domain fine-tuning please refer to section~\ref{appendix:expert_model_resources} in the Appendix.

\subsection{Datasets}

All the datasets listed here are widely used by LLM developers~\citep{touvron2023llama,touvron2023llama2,jiang2023mistral} to evaluate model performance in commonsense reasoning, coding, and medical domains. To generate the final training and testing data for the investigating routing methods, we gather all records together from all datasets and perform a stratified 80\% train, 20\% test split per dataset.

\textbf{Ai2-ARC~\citep{clark2018arc}.} The Ai2-ARC dataset consists of 7,787 natural science questions designed for standardized tests. We use its challenge partition with 2,590 samples, which includes only those questions that were answered incorrectly by both a retrieval-based algorithm and a word co-occurrence algorithm. 

\textbf{GSM8k~\citep{cobbe2021gsm8k}.} GSM8k is a high-quality dataset of grade school-level math word problems, covering relatively simple math concepts with 7,473 training and 1,319 testing samples.

\textbf{MBPP~\citep{austin2021mbpp}.} The MBPP dataset contains 974 basic programming problems suitable for entry-level programmers. It also includes text descriptions of the problems and test cases for functional correctness verification.

\textbf{PubMedQA~\citep{jin2019pubmedqa}.} The PubMedQA dataset is a biomedical question-answering dataset designed for answering research questions with yes/no/maybe responses. It contains 1,000 manually labeled question-answer pairs for cross-validation and testing.

\subsection{Methods}\label{subsec:routing_methods}
Below, we describe the predictive routing methods we used during our evaluation.

\textbf{Random-Router.} To evaluate the performance of a random router, for every test query we randomly pick an expert to execute the query. After performing this step for all test queries, we repeat the entire process for 10 times. Let \(\mathbf{E} = (e_1, e_2, \ldots, e_N)\) be the collection of all experts, we randomly select an expert from \(\mathbf{E}\) in each trial. Let \(e_i^j\) denote the $i$ expert randomly selected in the \(j\)-th trial, then the entire random expect selection process can be represented as: \(\{e_i^1, \ldots, e_i^{10}\}\). Once the collection of random experts is assembled, we submit the test query to each expert and collect all measurements to compute the evaluation metrics.

\textbf{kNN-Router.} The kNN-Router first encodes all training queries $\mathbf{q_i} \in D^t$ using a sentence transformer. Then, for every test query, $\mathbf{q_t}$, it finds its closest training query $\mathbf{q_i^\prime}$ in terms of cosine similarity in the embedding space and subsequently executes the test query using the expert that exhibited the best performance for the most relevant training query. The best performing expert $\mathbf{e_i^\prime}$ is the expert whose BERTSim score is the highest out of all the training query's experts, $q_i^\prime(E)$: 
$$\mathbf{q_i^\prime} = \min_{i \in D^t}(\frac{\mathbf{q_i} \cdot \mathbf{q_t}}{\|\mathbf{q_i}\| \|\mathbf{q_t}\|})$$
$$\mathbf{e_i^\prime} = \max_{j \in q_i^\prime(E)}(BERTSim_j)$$
A schematic flow of the 1NN-Router's embedding similarity and expert selection is also shown in Figure~\ref{fig:nn_router_pipeline}. Given that we only need to find the most similar training query to a given test query, we subsequently refer to this method as \textit{1NN-Router}. 

\textbf{MLP-Router.} For learning our predictive MLP-Router, we use a simple 2-layer perceptron:
$$
y_k = \phi \left( \sum_{j=1}^{m} w_{jk}^{(2)} \sigma \left( \sum_{i=1}^{n} w_{ij}^{(1)} x_i + b_j^{(1)} \right) + b_k^{(2)} \right)
$$
To train the MLP model, we convert the training queries into their vector representation by fitting a Bag-of-Words model. To learn the ranking of experts in terms of prediction performance, we use cross entropy loss on the scaled BERTSim scores. We used ReLU $(\sigma)$ and softmax $(\phi)$ as the hidden and output layers' activation function, respectively.

\textbf{BERT-Router.} To learn the BERT-Router, we performed a full parameter finetuning on a BERT model for sequence classification. We appended  a classification head with a softmax activation funciton on top of BERT's final hidden layer outputs to map the BERT embeddings $H$ to the desired number of experts (classes):
\[
y = \text{softmax}(W H + b), \; H = \text{BERT}(X)
\]
To fine-tune BERT, we first tokenize and encode all input training queries' text sequences $X$ using the BERT tokenizer and then update the pre-trained BERT model weights for a small number of epochs using cross entropy loss. Similar to the MLP-Router model, we train BERT-Router using the soft labels created by the scaled BERTSim scores.  

\textbf{Zero-Router.} Following the work of~\cite{hu2024routerbench}, we also evaluate the performance of the routing methods against the average performance of the available LLMs without any routing logic (lower bound), i.e., no-routing approach.

\textbf{Optimal.} We compare against two optimal cases (upper bounds), one refers to the optimal BERTSim performance per dataset (shown in Figure~\ref{fig:router_dataset_performance_bertsim}), and the other to the optimal performance recorded across all three evaluating dimensions (i.e., cost, throughput, model performance, shown in Figure~\ref{fig:optimization_trilemma}). In the former case, the optimal value is measured by averaging the best BERTSim score recorded for every test query by any expert. In the latter case, the optimal set of values is the minimum cost, maximum throughput and maximum performance recorded by any expert model or router method.

\subsection{Evaluation}
\begin{figure}[htpb]
    \centering    
        \centering
        \includegraphics[width=0.4\linewidth]{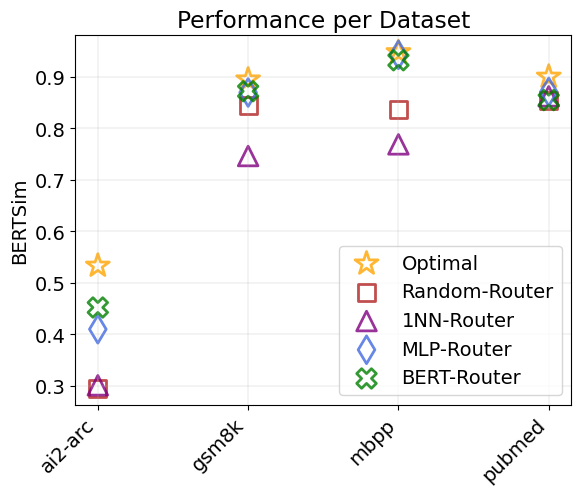}
        \caption{Router performance per dataset: BERT similarity score.}
        \label{fig:router_dataset_performance_bertsim}
\end{figure}

\begin{figure}[htpb]
        \centering
        \includegraphics[width=0.4\linewidth]{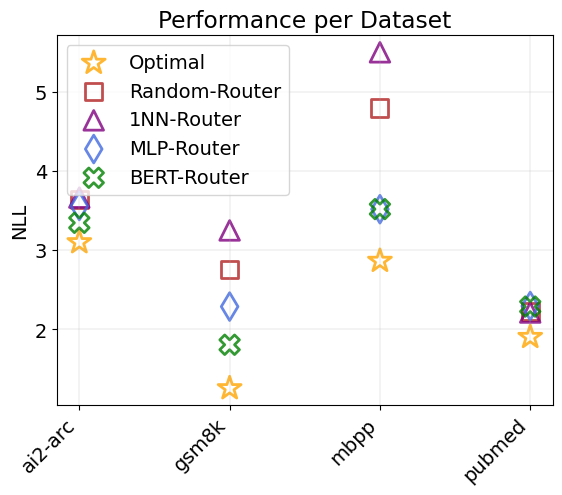}
        \caption{Router performance per dataset: Negative Log-Likelihood.}
        \label{fig:router_dataset_performance_nll}
\end{figure}

To systematically evaluate all investigating expert models in terms of query response times, we deployed each model on a machine employed with 8 NVIDIA DGX H100 GPUs. Figures~\ref{fig:router_dataset_performance_bertsim} and~\ref{fig:router_dataset_performance_nll} show the BERTSim score and NLL value comparison between all routing and optimal methods.

From the router vs. router comparison in Figures~\ref{fig:router_dataset_performance_bertsim} and~\ref{fig:router_dataset_performance_nll}, it is shown that naive methods, such as Random-Router or 1NN-Router that do not learn the embedding space can lead to suboptimal performance, cf. 0.3 BERTSim score for Random- and 1NN- Routers to 0.4 and 0.45 of MLP- and BERT- Routers in the Ai2-ARC dataset. Analogously, when it comes to train routing models that learn the embedding space, cf. BERT-Router to MLP-Router, more complex routing methods (i.e., BERT-Router) can lead to better outcomes and match closer the optimal performance, especially in challenging domains like GSM8K, cf. BERT-Router's NLL value of 1.803 to MLP-Router's 2.286.

To conduct a more thorough evaluation between expert models and routing methods, in Table~\ref{tbl:recorded_values_comparison}, we record all the numerical values collected throughout our experiments in terms of total monetary cost, query throughput, BERTSim score and NLL value. For every evaluating dimension, we also highlight with different colors the top-3 positions/rankings. The recorded values for the Zero-Router and the Optimal across all four dimensions are, Zero-Router: \{\$0.161, 153.242, 0.707, 3.295\} and Optimal: \{\$0.118, 214.925, 0.783, 2.326\}; we do not report these values in the table to emphasize the ranking between routing methods and standalone models. The MistralAI-7B exhibits the worst performance across all expert models, while the more recent small language model, Fox-1.6B, has the best performance across all expert models and evaluating dimensions. However, independent of Fox's performance, the collective model power provided by the routing methods, especially of the BERT-Router method, outperforms any other standalone expert model.

\begin{table}[]
\centering
\begin{tabular}{@{}lcccc@{}}
\toprule
\textbf{Model / Router} & \textbf{Total Cost}                                             & \textbf{Throughput}                      & \textbf{BERTSim}                    & \textbf{NLL}                           \\ \midrule
BioLlama-8B             & \$0.195                                                         & 155.613                                  & 0.686                                  & 3.408                                  \\
BioMistral-8B           & \cellcolor[HTML]{FFCCC9}\textbf{\$0.125}                        & 208.399                                  & 0.669                                  & 3.581                                  \\
CodeLlama-7B            & \$0.156                                                         & 102.993                                  & 0.694                                  & 3.299                                  \\
Fox-1.6B                & \cellcolor[HTML]{8AD871}{\color[HTML]{000000} \textbf{\$0.118}} & \cellcolor[HTML]{8AD871}\textbf{214.925} & \cellcolor[HTML]{FFCCC9}\textbf{0.761} & \cellcolor[HTML]{A5D8E7}\textbf{2.958} \\
MathDeepSeek-7B         & \$0.138                                                         & 187.166                                  & 0.746                                  & 3.286                                  \\
MistralAI-7B            & \$0.223                                                         & 89.587                                   & 0.694                                  & 4.205                                  \\
Qwen-7B                 & \$0.164                                                         & 114.008                                  & 0.698                                  & \cellcolor[HTML]{8AD871}\textbf{2.326} \\ \midrule
Random-Router           & \$0.143                                                         & \cellcolor[HTML]{FFCCC9}\textbf{209.171} & 0.715                                  & 3.316                                  \\
1NN-Router              & \$0.131                                                         & 205.715                                  & 0.697                                  & 3.271                                  \\
MLP-Router              & \$0.147                                                         & 177.508                                  & \cellcolor[HTML]{A5D8E7}\textbf{0.773} & 3.164                                  \\
BERT-Router             & \cellcolor[HTML]{A5D8E7}\textbf{\$0.122}                        & \cellcolor[HTML]{A5D8E7}\textbf{213.145} & \cellcolor[HTML]{8AD871}\textbf{0.783} & \cellcolor[HTML]{FFCCC9}\textbf{3.091} \\ \bottomrule
\end{tabular}
\caption{Total querying cost, mean throughput and cosine similarity between predicted and expected answers per model and router considering all the four benchmark datasets. Box coloring represents the following ranking column-wise: \colorbox[HTML]{8AD871}{rank 1}, \colorbox[HTML]{A5D8E7}{rank 2}, \colorbox[HTML]{FFCCC9}{rank 3}.}
\label{tbl:recorded_values_comparison}
\end{table}

\begin{figure}[htpb]
    \centering
    \includegraphics[width=0.5\linewidth]{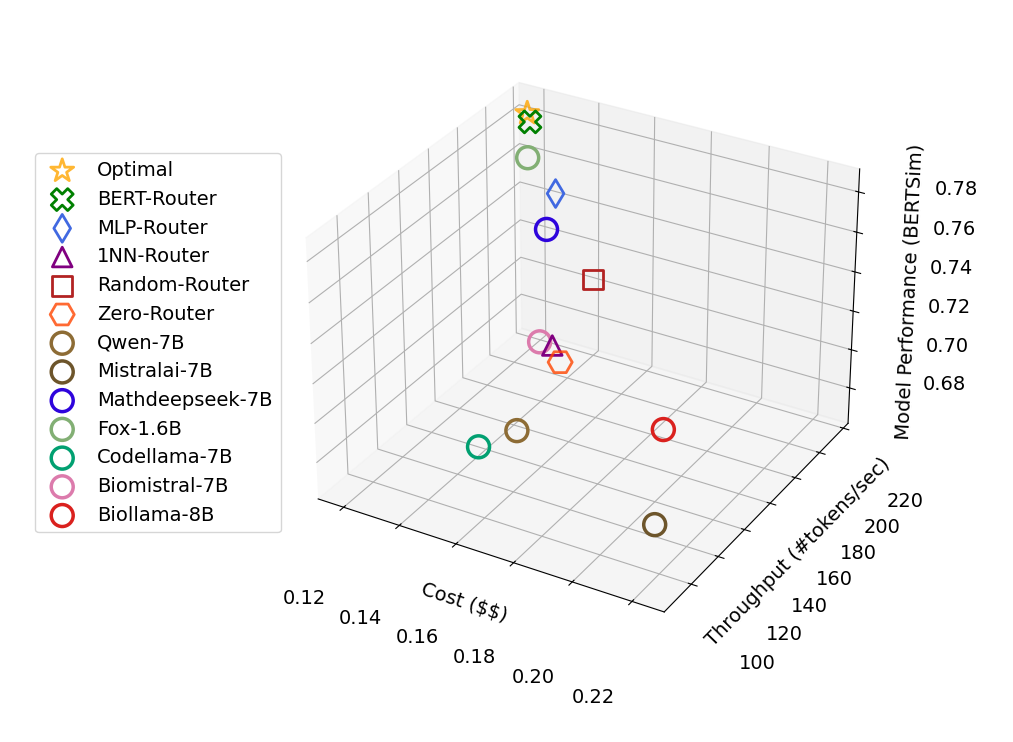}
    \caption{A holistic view of model performance, throughput and total querying cost for standalone deployed expert models and different routing methods.}
    \label{fig:optimization_trilemma}
\end{figure}

By using as a reference routing method the BERT-Router approach and baseline the mean performance of all standalone model experts (i.e., the Zero-Router), we find that the BERT-Router leads to a close of 30\% cost reduction and 40\% query inference throughput increase compared to no routing at all. At the same time though, BERT-Router is capable of maintaining or slightly enhancing the average mean model performance, by a 11\% in terms of BERT similarity score and lead to a 6\% NLL reduction.

Finally, in Figure~\ref{fig:optimization_trilemma}, we provide a 3D visualization of the optimization trilemma problem w.r.t. total monetary cost (x-axis), query throughput (y-axis) and model performance (z-axis). The Figure clearly shows that the BERT-Router method outperforms all other expert models and routing methods across all three evaluation criteria, while almost matching the optimal performance.

\section{Unlock the Potential of Collaborative Routing: Edge-to-Cloud}
\begin{figure}[htpb]
    \centering
    \includegraphics[width=0.7\linewidth]{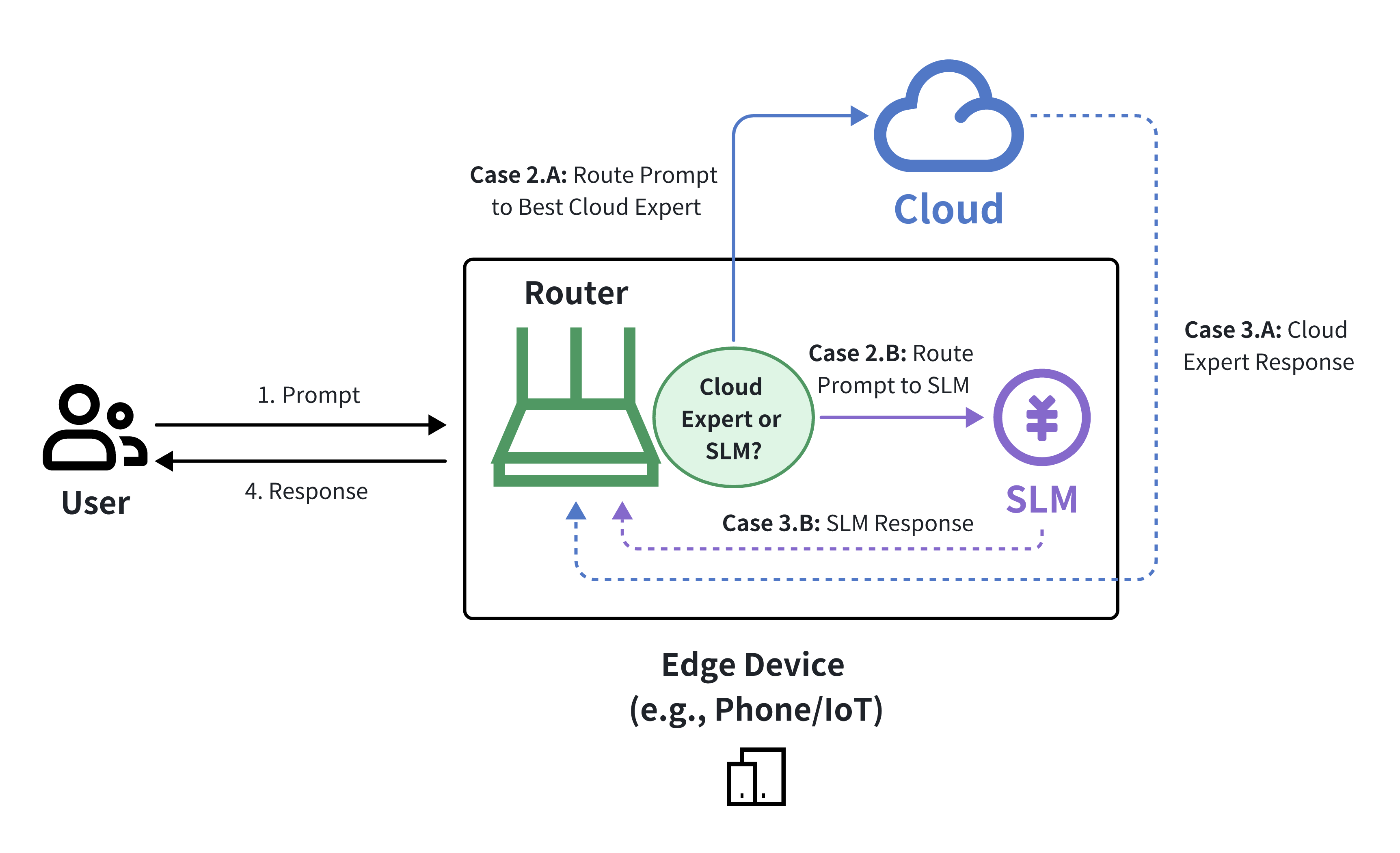}
    \caption{Answering queries locally on the edge through an SLM or proxying to the cloud.}
    \label{fig:small_to_big}
\end{figure}

So far, we have discussed routing methods that can route query prompts to the most suitable expert hosted on a cloud service. However, the bigger promise of a model router is how to be effectively deployed on edge devices to help decide on whether a query prompt can be answered locally by a small model running on the edge or routed to an expert hosted on the cloud. Figure~\ref{fig:small_to_big} shows such an architecture, where a model router is deployed locally on the edge device. The router is responsible for deciding whether a user's new request submitted to the edge device should be answered directly by a small model, such as a small language model (SLM), which is already running on the edge, or by a larger model deployed on a cloud provider, or through a combination of the two. By applying this approach, we can significantly reduce querying and communication costs on the edge while maintaining overall model and query performance. 

\begin{figure}[htpb]
    \centering
    \includegraphics[width=0.45\linewidth]{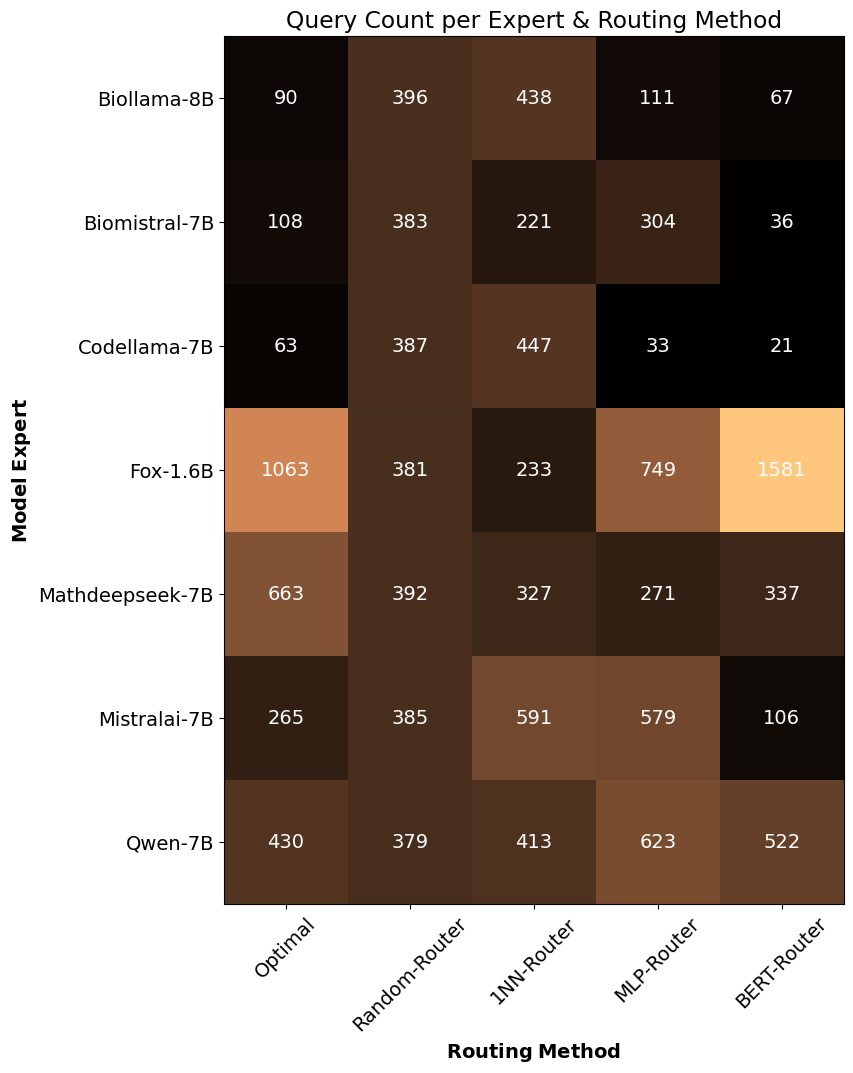}
    \caption{Number of test queries allocated to each model expert by each routing method.}
    \label{fig:query_count}
\end{figure}

To provide further insights towards the materialization of this edge-to-cloud collaborative approach, in the heatmap shown in Figure~\ref{fig:query_count}, we record the number of test queries answered by each expert model for every routing method. From the reported values, it is apparent that both the MLP-Router and the BERT-Router, route most of the test queries investigated during our experimental analysis (section~\ref{sec:experiments})  to the Fox-1.6B small language model, which is similar to the behavior observed by the Optimal (oracle) approach. However, other approaches like the Random-Router and 1NN-Router, distribute almost equally the number of queries across all model experts. 

If we assume that an SLM, like the Fox-1.6B model, is deployed on the edge, our analysis shows that by learning the embedding space of existing query prompts using an MLP or BERT-based routing approach, the majority of queries will be forwarded to the most suitable expert, which in this case is the SLM. As a result, deploying such routing methods on the edge enables edge devices to decide whether a query prompt should be routed to the local SLM or a cloud model expert, paving the way for the effective development of collaborative edge-to-cloud model routing techniques.

\section{Conclusion}

We present our multi-LLM routing system, called \textit{TO-Router}, for the first time. With the TO-Router system, users can easily interact with multiple LLM expert models hosted on the same or different platform providers, without being limited to a single monolithic LLM system. By utilizing a routing method capable of learning the embedding space of query prompts, such as MLP- and BERT-based methods, users can benefit from significant cost savings (up to 30\%), improved query response times (up to 40\%), and enhanced model performance (up to 10\%). As part of our immediate future plans, we aim to evaluate the feasibility of dynamically adding and removing model experts during the router's endpoint deployment and test the routing efficacy of both small and large pre-trained language models.

\bibliographystyle{plainnat}
\bibliography{ref}

\clearpage
\appendix
\section{kNN-Router Diagram}

\begin{figure}[htpb]
    \centering
    \includegraphics[width=\linewidth]{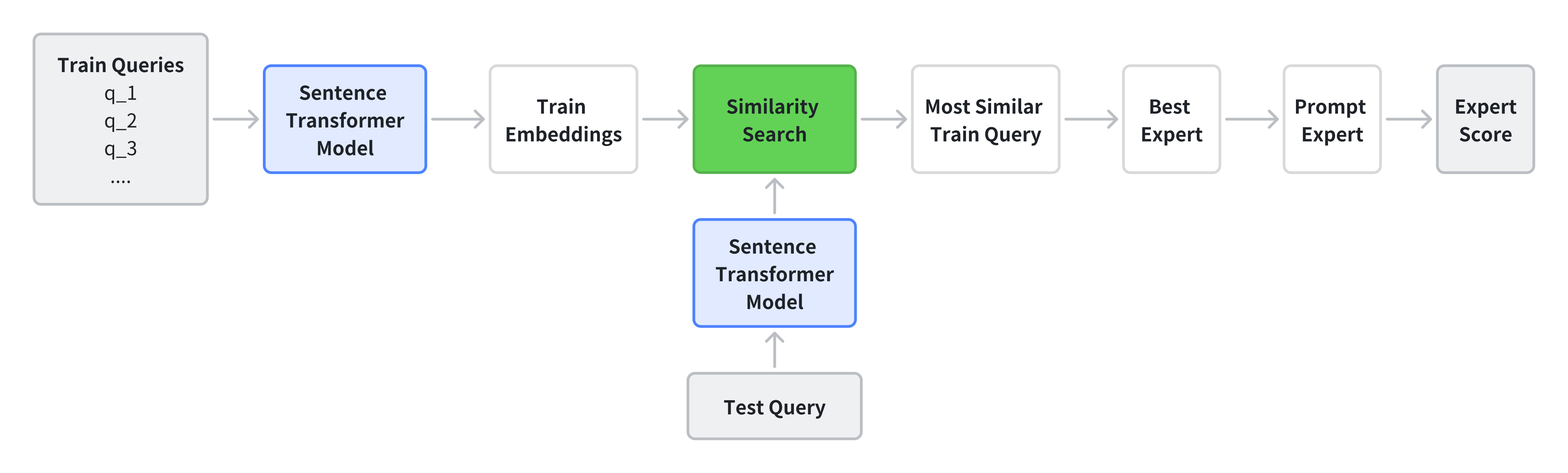}
    \caption{A flow diagram of the embedding similarity approach used by the 1NN-Router.}
    \label{fig:nn_router_pipeline}
\end{figure}

\section{Model Pricing}

\begin{table}[htpb]
\centering
\scriptsize
\begin{tabular}{@{}lcc@{}}
\toprule
Model Type  & \$\$ / 1M Input Tokens & \$\$ / 1M Output Tokens \\ \midrule
DeepSeek-8B & \$0.14                 & \$0.28                  \\
Fox-1.6B    & \$0.20                  & \$0.20                   \\
Llama-8B    & \$0.20                  & \$0.20                   \\
Mistral-8B  & \$0.25                 & \$0.25                  \\
Qwen-7B     & \$0.20                  & \$0.20                   \\ \bottomrule
\end{tabular}
\caption{Price per million input and output tokens for different types of model architectures.}
\label{tbl:model_pricing}
\end{table}

\section{Router Models Data Preparation}
To generate experts' soft labels to train the MLP and BERT-Router models, we used the BERT similarity scores and set the temperature value of the softmax function to 10, i.e., $T = 10$. To compute the closest training query to a given test query in the case of the 1NN-Router, we compute the queries' embeddings using the sentence transformer library.~\footnote{\url{https://www.sbert.net/docs/quickstart.html}}

\section{Router Models Training Hyperparameters}
The total number of experts is 7. The MLP-Router's hidden layer size is 256. The random seed for all experiments is set to 42. The applied optimizer for training both the MLP and BERT routers is Adam with weight decay, the learning rate is set to $5e-3$ and $5e-5$, respectively. We also applied L2 norm regularization with $\lambda=1e-4$. The batch size is set to 8 and the total number of training (MLP model) and fine-tuning (BERT model) is set to 5 epochs. The BERT model for the router is \textit{bert-base-uncased}. To counter dataset class/expert imbalance we observed while generating the training and testing datasets, i.e., an expert model might be more suitable to answer many more queries than other experts, we used a sample weighting function, with the weight of each sample being the inverse proportion count of samples per class in the entire training dataset, i.e., the total weight sample proportion for each class/expert $i$ across all experts $E$, is measured as $w_i = \tfrac{\sum_{j \in E}|D_j|}{|D_i|}, \forall i \in E$, with the final weight value per training sample being equal to $w_i = \tfrac{w_i}{\sum_{j \in E}|w_j|} \forall i \in E$.

\section{Expert Model Resources}\label{appendix:expert_model_resources}
Below, we provide details regarding the internal architecture and type of models we used as our expert models in this study. For every instructed model, if not otherwise specified, we set the maximum tokens generation length to 512, the temperature to 0.7, and the top-p parameter to 0.95.

\begin{itemize}
    \item \textbf{BioLlama-7B}~\footnote{\url{https://huggingface.co/aaditya/Llama3-OpenBioLLM-8B}}:This model is an advanced Llama-3-based model designed specifically for the biomedical domain. With policy optimization and a custom medical instruction dataset, it outperforms even the ChatGPT API. Following the recommended parameters, we set max new tokens to 256, temperature to 0.1 and top-p to 0.9.
    \item \textbf{BioMistral-7B}~\footnote{\url{https://huggingface.co/BioMistral/BioMistral-7B}}: This Mistral-based model, pre-trained using textual data from PubMed Central Open Access, is well-suited for medical domains and achieves performance comparable to the ChatGPT API across all medical evaluation benchmarks.
    \item \textbf{CodeLlama-7B}~\footnote{\url{https://huggingface.co/codellama/CodeLlama-7b-hf}}: This model adapts the Llama-2-7B model with a large collection of code datasets, incorporating an infilling training objective and long input context subsets.
    \item \textbf{Fox-1.6B}~\footnote{\url{https://huggingface.co/tensoropera/Fox-1-1.6B-Instruct-v0.1}}: Fox-1 is a decoder-only transformer-based small language model with 1.6B parameters, developed by TensorOpera AI. Fox-1-Instruct-v0.1 is an instruction-tuned version with an 8K native context length, finetuned with 5B tokens of instruction-following and multi-turn conversation data.
    \item \textbf{Mistral-7B-Instruct}~\footnote{\url{https://huggingface.co/mistralai/Mistral-7B-Instruct-v0.2}}: This model is an officially released instruct fine-tuned version of the Mistral-7B-v0.2.
    \item \textbf{Qwen-7B-Instruct}~\footnote{\url{https://huggingface.co/Qwen/Qwen2-7B-Instruct}}: This model is an officially released instruct fine-tuned version of the Qwen2-7B.
    \item \textbf{MathDeepSeek-7B}~\footnote{\url{https://huggingface.co/deepseek-ai/deepseek-math-7b-instruct}}: This model, initialized with DeepSeek-Coder-v1.5 7B, continues pre-training on math-related tokens sourced from the web, achieving impressive scores on the competition-level MATH benchmark. Following the recommended parameters, we set max new tokens to 512, top-k to 50 and top-p to 0.95.
\end{itemize}

\end{document}